\documentclass[10pt,twocolumn,letterpaper]{article}

\usepackage{btas}
\usepackage{times}
\usepackage{epsfig}
\usepackage{graphicx}
\usepackage{amsmath}
\usepackage{amssymb}
\PassOptionsToPackage{hyphens}{url}\usepackage{hyperref}

\btasfinalcopy 

\ifbtasfinal\fi

\begin{document}

\title{Four Principles of Explainable AI as Applied to Biometrics and Facial Forensic Algorithms}

\author{P. Jonathon Phillips  \hspace{0.5in} Mark Przybocki\\ \\
National Institute of Standards and Technology\\
100 Bureau Dr, Gaithersburg, MD 20899, USA\\
{\tt\small jonathon.phillips@nist.gov, mark.przybocki@nist.gov}
}

\maketitle
\thispagestyle{empty}

\begin{abstract}
Traditionally, researchers in automatic face recognition and  biometric technologies have focused on developing accurate algorithms. With this technology being integrated into operational systems, engineers and scientists are being asked, do these systems meet societal norms? The origin of this line of inquiry is `trust' of artificial intelligence (AI) systems. In this paper, we concentrate on adapting explainable AI to face recognition and biometrics, and we present four principles of explainable AI to face recognition and biometrics. The principles are illustrated by \emph{four} case studies, which show the challenges and issues in developing algorithms that can produce explanations.  
\end{abstract}

\section{Introduction}

In operational settings, face recognition and biometric systems answer questions about a person's identity. Typical questions include ``Are the faces in two images the same person?" and ``Retrieve the top 10 candidates from a search." Based on the answer, a decision is made, and it influences a resulting action by either a person or a machine. An example is unlocking a biometric-enabled smartphone. The phone accepts that the biometric belongs to an approved user and unlocks the phone. When the phone does not accept the biometric verification attempt, the user has the option to enter a passcode.  The process for accessing the phone is simple and straightforward.

For high-stakes scenarios, identity decisions require a justification that supports the AI finding. When facial forensic examiners testify in court, they explain {\it how} they came to their conclusions.  As face recognition systems increase in accuracy, forensic analysis will include decisions by machines, and in testimony, forensic examiners will need to explain {\it how} the algorithm contributed to the conclusion.

As biometrics continue to assist humans in making high-stakes decisions, society will view these systems in how they meet societal norms. For example, one area of concern is the impact of bias in face recognition \cite{Phillips:2010fk,Lui:2009fv,buolamwini2018gender,Krishnapriya2018}.

Failure for a system to articulate the rationale for a negative answer will affect the level of trust users will grant a system. It will sow suspicion that the system is inherently biased or unfair. It will slow acceptance by society and the adoption of the technology may ultimately fail. Humans often accept or reject innovations and new technology based on their gut feeling.

Explainable AI is one of several properties that characterize "trust" in AI systems.  These properties include fairness, bias, transparency, and accountability among others. The definitions of these properties vary by author. In most cases, these terms are not defined in isolation, but as a part or set of principles or pillars. The focus of the principles is not algorithmic methods, such as deep convolutions neural networks, but societal norm of AI systems. While we acknowledge that a trade-off can exist between an AI system's accuracy and its explainability, this article focuses on properties that AI systems should exhibit as functioning systems in society. These principles may affect the methods in which algorithm operate so that they meet explainable goals.  These principles may form the foundation of policy considerations, safety, acceptance by society, ethics, and other aspects of AI technology.

There are many valid principles and concepts related to explainable AI. Some of which are necessary to provide balance by domain, or provide special consideration by certain AI interest groups (researcher, user, policy maker, etc.).

Our approach is to focus on the core concepts of explainable AI. We treat explainability as a distinct concept in AI, while at the same time acknowledging that other AI concepts are necessary to enhance the characterization of a functioning system in society. The result of adopting these principles will assist in guiding research direction and helping to define best practices for developers. Principles of explainable AI will form one of the bases of addressing fairness, bias, transparency, security, safety and ultimately trust in AI systems.

We characterize explainable AI as applied to face recognition and biometrics by four principles. The face recognition case studies presented include biometric, forensics and general applications.   The \textbf{four contributions} of this paper are
\begin{itemize}
    \item Articulate four principles of {\it explainable AI} applied to Biometrics and Facial Forensic Algorithms. 
    \item Describe how the four principles apply.
    \item Through \textbf{four} case studies, review challenges in implementing explainable AI. 
    \item Examine the relationship between explanations and response time to decisions.  
\end{itemize}
\vspace{0.5 cm}

After describing the four principles, we will illustrate the principles with four case studies in face recognition and biometrics. The case studies will cover challenges and the impact of adopting the principles. 

\section{Four Principles of Explainable AI}
\label{sec:FourPrinciples}

Our four principles of explainable AI for face recognition and biometrics are

\begin{description}
    \item[Explanation:] Systems deliver accompanying evidence or reason for all decisions.
    \item[Interpretable:] Systems provide explanations that are understandable and meaningful to individual users. 
    \item[Explanation Accuracy:] 
        Systems provide explanations that are understandable, meaningful, and acceptable to individual users.
    \item[Knowledge Limits:] The system only produces decisions under the conditions it was designed and tested.
\end{description}

\subsection{Explanation}

The {\it Explanation} principle obligates systems to supply evidence, support or reasoning for each decision.  By itself, this principle does not require that the evidence be correct, informative, or even make sense; it merely states that a system is capable of providing an explanation.

\subsection{Interpretable}

A system fulfills the {\it Interpretable} principle if the user understands the system's explanations and the explanation allows users to complete their tasks. This principle does not expect a one-size fits all explanation. There can be multiple groups of users for a system, and the Interpretable principle dictates that explanations are tailored to each of the user groups.  Thus, a system may need to generate multiple explanations. We will refer to this as a \emph{multi-sided} explanation. 

The explanations should be generated when a decision is computed; however, the explanations may not be consumed contemporaneous with a decision.  For corporate smartphones, face recognition unlocks the phone, and an explanation is not provided to the user. The explanation is stored for subsequent security audits by the company.

\subsection{Explanation Accuracy}
\label{sec:Accuracy}

Together, the Explanation and Interpretable principles only obligate a system to produce explanations that are understandable by a user community.  These two principles do not mandate that a system delivers accurate explanations.  The {\it Explanation Accuracy} principle imposes accuracy on a system's explanations.

For over 25 years, researchers in face recognition and biometrics have measured the accuracy of algorithms and systems decision on benchmark data sets \cite{PhillipsMoonRizviRauss2000,Phillips:2009zl,Phillips:2008eu,przybocki2007nist,sadjadi20172016}. To avoid confusion, we will refer to this as \emph{decision accuracy}. Standard decision accuracy statistics include the verification and false accept rates for an operating point and the rank one retrieval rate for searching a gallery.  

Decisions and explanation accuracies are independent. Regardless of whether the system decision is ultimately correct or incorrect, an explanation must describe how the system came to its conclusion.  

An accurate explanation does not imply that a system provided the correct answer. If the system decision is wrong, the explanation needs to articulate how the system came to that wrong decision.

For correct answers, there are two reasons an explanation could be wrong.  In the first case, the explanation is not plausible. For example, the object on the road is a car because it has two legs and three wings.  In the second case, the explanation is plausible but does not describe the reasoning the system used. For example, the explanation provided states that is a car because it is the right size and has four wheels. The correct logic is that the object looks like a model of car in the training set.

While the community has established decision accuracy metrics, researchers have not developed performance metrics for explanations. 

\subsection{Knowledge Limits}

Given a pair of non-frontal face images, a face recognition system can provide a judgment for the similarity between the two faces.  If engineers designed the system to process frontal faces, then the system should be prohibited from comparing these two non-frontal images. The {\it Knowledge Limits} principle commands systems to flag cases the algorithm was not designed or tested to solve.   The Knowledge Limits principle differs from the Explanation Accuracy principle in that it doesn't require evidence of truth, instead it provides a mechanism for the system to punt on the answer.  In the above example, the algorithm should not compare the two non-frontal faces and alert the system operators.

\section{Related Work}
\label{sec:RelatedWork}

Others have articulated the pillars of explainable AI \cite{ExplainableModelsforHealthcareAI}, \cite{dignum2017responsible},\cite{guidotti2018survey}.  These pillars describe explainable AI's relationship to transparency, trust, fairness, and other properties.  The focus of our principles was independent from these other terms. Guidotti et al.\cite{guidotti2018survey} provide a detailed survey of methods for explaining black-box algorithms.

The Defense Advanced Research Projects Agency (DARPA) runs the Explainable Artificial Intelligence (XAI) Program which is designed to produce more explainable models, while maintaining a high level of learning performance (prediction accuracy); and to enable human users to understand, appropriately trust, and effectively manage the emerging generation of artificially intelligent partners\footnote{\url{https://www.darpa.mil/program/explainable-artificial-intelligence}}.

\section{Dance Floor Perspective}

\begin{figure*}[ht!]
\begin{center}
\includegraphics[width=5in]{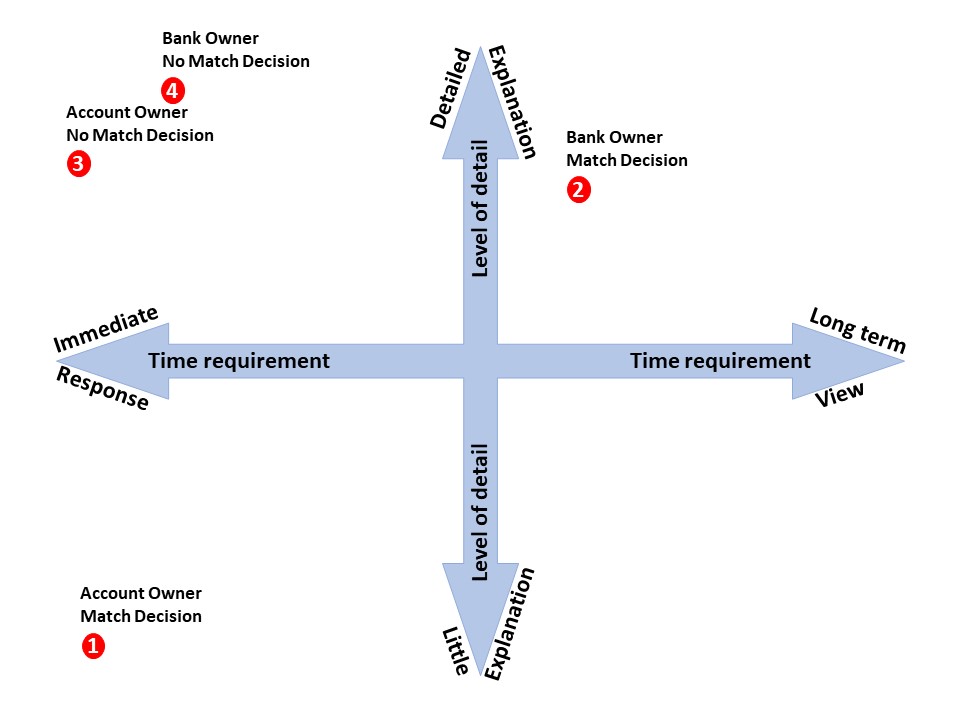}
\caption{Dance floor that shows length of response time versus distance from action.}
\label{fig:DanceFloorDec18}
\end{center}
\end{figure*}

We characterize interpretability along two axes, see Figure~\ref{fig:DanceFloorDec18}. In Figure~\ref{fig:DanceFloorDec18}, the horizontal axis represents the {\it time requirement} a user has to respond to a situation. The time requirement axis addresses situations from those that require immediate responses to those that permit a long term view.  The vertical axis represents the {\it user requirement} for how the system information will be used. The user requirement axis addresses situations related to the level of detail the consumer or user will require, from little explanation needed or required to much greater detail being required. 

What is interpretable depends on the consumer of the results. Ability to explain depends on the technique or the domain. Consider how decisions from AI systems might be used in the situations presented in the four quadrants of Figure~\ref{fig:DanceFloorDec18}. The LAW OFFICER (lower-left quadrant) in the context of relying on facial recognition in an active crime situation may use AI decisions to determine if a suspect should be brought in for questioning.  In this case, the explanation need that is meaningful to her is minimal, and simply supports why the system yields a match/unmatched decision. The DETECTIVE (lower-right quadrant) in the context of building a case may require the AI 'match' decision to include a greater explanation that will support arresting an individual.  The JURY (upper-left quadrant) may need the AI system to explain greater details in possibly more layperson terms about the performance of the system under similar and dissimilar situations. And the CONVICT (upper-right quadrant) may want to dissect the most minute details of the system match finding, as she prepares for her appeal.


\section{Case Studies}

This section presents four case studies that explore the challenges and identify research directions of explainability in face recognition and biometrics.   

\subsection{Explainable Methods for DCNN-based Algorithms}

Our first case study looks at deep convolutional neural networks (DCNN)-based algorithms for visual biometrics and highlights three of the challenges in explainable AI. First, what is an explanation? Second, does a user understand the explanation? Third, does the explanation accurately reflect the DCNN decision process?

\begin{figure}[ht!]
\begin{center}
\includegraphics[width=2.75in]{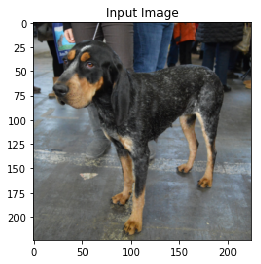} \\
\includegraphics[width=2.75in]{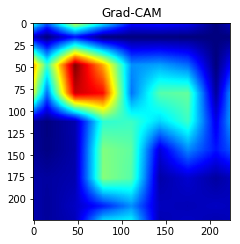}
\caption{Example of the gradCAM technique~\cite{selvaraju2017grad}. The top image is the input to the gradCAM algorithm and the bottom images is a heatmap from analyzing the original image. The highlighted regions in the bottom images shows the regions that contributed to the objects being classifying as a dog.  }
\label{fig:CAMeexample}
\end{center}
\end{figure}

In daily human communications, the response to a request for an explanation is usually spoken or written.  Visual explanations are reserved for special conditions, and they are often accompanied by verbal clarification.  Two of the most popular explanatory techniques, class activation mapping (CAM) \cite{zhou2016learning} and grad-CAM \cite{selvaraju2017grad}, provide pure visual explanations, see Figure~\ref{fig:CAMeexample}.  In Figure~\ref{fig:CAMeexample}, the classifier reports that there is a dog in the image and the explanation of the decision is the highlighted region in the image. Since the output of the gradCAM algorithm is the heatmap, further interpretation of the visual explanation is left to the consumer of the image and its highlighted region.

The CAM and grad-CAM methods provide coarse localization of regions in an image.  In visual biometrics, explanations will require more precise explanations, see the facial forensic case study in Section~\ref{sec:CaseStudyForensicFace}.

The technique in Hendricks et al.~\cite{hendricks2016generating} provides a written explanation for the decision of a bird species classifier. The algorithm learns explanations from expert birders.  The experiments in the study measure the explanation accuracy by having human birders judge the algorithm's explanations.

The CAM and grad-CAM methods demonstrate there are different styles for providing explanations, from purely verbal or visual to a combination of both.  The evaluation protocol in Hendricks et al. measured explanation accuracy. The existing literature gives insight to the first two questions raised at the beginning of this case study.  Unfortunately, the third question was not explicitly addressed. 

\subsection{Speaker Recognition}

Speaker recognition technology at its core compares two samples of speech and determines the likelihood that the same speaker is speaking in each. If the system-dependent internal score is above a predefined threshold, a match is declared. Current systems are not required to explain the decision they produce, although in NIST Speaker Recognition Evaluation 2018 some neural network models incorporated "attention" mechanisms to help explain which features the model was focusing on to produce certain outputs \cite{wang2018attention}. Speaker recognition technology is used as a biometric (e.g., for access to a room or a bank account), for investigatory work (e.g., does a suspect's voice match that of a voice on a 911 call), or for search (e.g., to sift through voice mails looking for a particular message).

Consider the situation of a person accessing his bank account using speaker recognition technology. The system will compare the voice sample collected over the phone to a signature file stored in its database and respond with either TRUE or FALSE, depending on whether it decides the voice samples match.

Placing this example in the context of explainable AI, the Explanation Principle requires the system to add reasoning to each decision. The principle does not define the content or the quality of the reasoning and therefore meeting this principle in practice should be straightforward. The reasoning or evidence may take many forms. It could be a confidence score, a log-likelihood score, or a data quality score; or it could take the form of a series of technical statistics. Maybe there are internal system rules which predicate a FALSE decision, such as having the number of attempts exceeded, the account is locked, or the system is unavailable or failure to connect to the server.  

The Interpretable Principle, on the other hand, does introduce challenges to explainable AI in the context of speaker recognition technology. The first challenge is in determining who the explanation is for. In our example, we have two possible consumers of the AI decision - the account owner and the bank owner. Once the consumer of the AI decision is identified, the second challenge is in creating an explanation with the appropriate level of detail. To demonstrate these challenges, we refer to the red dots in Figure~\ref{fig:DanceFloorDec18} showing that multiple consumers of the decision and explanation sit in various quadrants of the dance floor, requiring different levels of detail in the explanations. (1) Represents operation as expected from the account owner's point of view. Immediate correct response with very little explanation required. (2) Represents the same situation from the bank owner's point of view. The bank owner would not be checking every login attempt as it happens but may keep records to consult in case of reported fraudulent access.  Here the bank owner has less of a time requirement but a greater need for details in the explanation. (3) Represents a system failure. The account owner will want to understand why he was denied access to his account.  He will expect the same timely response but with much greater detail in the explanation. (4) represents the same system failure from the bank owner's point of view where they may require additional information in the explanation to address the failure, and the bank owner would like to resolve the issue promptly. 

In terms of the Explanation Accuracy Principle the explanation must be proven via measurement to represent the reasoning for the decision. For current state-of-the-art systems the Explanation Accuracy Principle presents new challenges to speaker recognition technology. It is not our goal to solve these challenges, but rather to identify the area of research that will bring explainable AI to speaker recognition technology. A single algorithm may have multiple accuracy measures depending on the resulting AI decision, or method employed to reach a decision. This is demonstrable by considering the differences of measurements between a system defined no-match, and a system that couldn't connect to the server. The same type of results, would be explained and measured with different techniques. 

The concepts covered in the Knowledge Limits Principle are active in the technology implementation today. The use of thresholds can be expanded to include no-decision ranges to account for instances when the data quality is not sound, the presence of speech samples are too few, or the expected speech is replaced with yelling, crying, grunting, laughter, etc.

\subsection{AI Superior to Humans}

For an explanation to be accepted by users, it needs to align with their knowledge, background, and intuition.  What happens when an explanation does not align with prior experience? With advances in AI technology, there are numerous reasons this will occur.  In this case study, we discuss four reasons our intuition could not align with an algorithms explanation.

First, our intuition for how we come to decisions is not accurate. For example, in face recognition, Rice et al.~\cite{Rice:2013aa} showed that people do not know which facial features are critical for recognizing individuals.  The challenge in this situation is developing techniques that enable humans to accept counter-intuitive explanations. 

Second, a system has capabilities superior to humans. There are several avenues for superior capabilities.  First, algorithms perform better than the majority of humans; e.g., face recognition~\cite{phillips2018face}.  Second, algorithms combined with humans yield a superior performance;  e.g., face recognition~\cite{phillips2018face} — the source of these capabilities maybe new analysis paths.

Third, AI systems can find entirely novel solutions outside usual human reasoning paths, experience, and intuition. For example, Chef Watson created entirely novel recipes \footnote{\url{https://www.ibm.com/blogs/watson/2016/01/chef-watson-has-arrived-and-is-ready-to-help-you-cook/}}. It is possible, that a system could explain its reasoning process, but people will need time to understand and comprehend the explanation. 

Fourth, explaining decision in biometric  {\it Intelligent Infrastructure}~\cite{JordanMedium2018}. Intelligent infrastructure system performs AI functions on a large scale. Two examples are determining identity in social media networks or nation-scale mugshot and fingerprint systems.  Whereas people can compare two face images, they cannot process millions of samples in a multi-biometric search.  

For these four scenarios, the user group that can understand the rationale for a decision may be restricted to a specially trained group with access to a customized set of tools.  An additional responsibility for the specially trained group could be to disseminate an explanation to the public and other user groups. 

\subsection{Facial Forensics Examiners}
\label{sec:CaseStudyForensicFace}

Forensic facial examiners perform detailed analysis of face images.  From their analysis, they prepare comprehensive reports and testify in court if required.  Examiners prepare reports explicitly stating their conclusions following best practices\cite{FISWG2012}.  Ideally, AI systems would prepare reports to follow the same best practices. Unfortunately, this is beyond current technology; however, developing an AI system that collaborates with examiners to write reports maybe within reach.

\section{Conclusion}
We introduced the principles of \emph{Explanation}, \emph{Interpretable}, \emph{Explanation Accuracy}, and \emph{Knowledge Limits}, to provide a solid foundation for explainability in face recognition and biometrics. Instead of focusing on technology, the principles turn attention to systems meeting societal norms. We illustrated this point of view with four case studies that provide lessons and research directions. 

First, explanations come in different varieties that include visualizations for image-based biometrics, verbal description, and a combination for both.  Evaluation techniques need to be developed for these different explanation modes.

Second, the authors failed to find a significant effort in the face recognition and biometrics literature that addresses one of the principles, let alone all four principles.  Related work in computer vision could provide a starting point for further research and point to paths for attacking all four points.

Third, researchers need to tackle the challenge of creating explanations for systems with extra-human capabilities.  Extra-human capabilities include systems that are more accurate than humans, systems that involve the fusing of human and machine decisions using intelligent infrastructure.  One solution is to develop tools that assist specially trained experts in understanding the reasoning behind systems' decisions.

With the increase in complexity of face recognition and biometric systems, the need for specialized tools and training should not be a surprise.  For many complex information processing systems, experts regularly educate the public, senior manages and policy advisers on the details on how they are working.

The lessons learned by setting the foundations of explainability in face recognition and biometrics will inform the broader AI community as it addresses explainability, transparency, fairness, and trust.

{\small
\bibliographystyle{ieee}
\bibliography{ExplainableAI,faces2}
}

\end{document}